\newcommand\SEKIusersusepackages
{
\usepackage{verbatim}
\usepackage{wrapfig}
\usepackage{amsmath,latexsym,url}
\usepackage{xspace}
\usepackage{stmaryrd}
}

\newcommand\SEKIREVIEWSTATUS{1}

\newcommand\SEKIREVIEWER
{Claus-Peter Wirth 
\\E-mail: {\tt wirth@logic.at}
\\WWW: \url{http://www.ags.uni-sb.de/~cp}
}

\newcommand\SEKIREFEREEPROCESS
{LPAR 2003, 
 10th International Conference on Logic for Programming,
 Artificial Intelligence, and Reasoning;
 full paper, 15 pp.
}

\newcommand\SEKIDOCUMENTCLASS{0}

\newcommand\SEKIYEAR{2009}

\newcommand\SEKINUMBEROFYEAR{02}

\newcommand\SEKITYPE{0}

\newcommand\SEKIPUBLISHEDTYPE{0}
\title{Quantified Multimodal Logics\\ in Simple Type Theory} 
\author{Christoph  Benzm\"uller and Lawrence C. Paulson}

\newcommand\edcorr[2]{#2}
\newcommand\neginpit[1]{\neg(#1)}
\newcommand\mnotinpit[1]{\mnot(#1)}

\input seki-deckblatt-3

\begin{document}
\makecover

\input{macros.mac}
\maketitle
\begin{abstract}
 We present a straightforward embedding of quantified multimodal logic in
  simple type theory and prove its soundness and completeness. Modal operators are replaced by quantification over a type of possible worlds.
  We present simple experiments, using existing higher-order theorem provers,   to demonstrate that the embedding allows automated proofs of statements in these logics, as well as meta properties of them.
\end{abstract}

\vfill\pagebreak

\section{Motivation} \label{sec1}
\label{Motivation}
  There are two approaches to automate reasoning in
  modal logics.
  The \emph{direct}
  approach~\cite{baldoni96framework,DBLP:journals/fuin/Nguyen03}
  develops specific calculi and tools for the task; the \emph{translational}
  approach~\cite{DBLP:conf/jelia/Nonnengart94,DBLP:conf/cade/Ohlbach88}
  transforms modal logic formulas into first-order logic and applies
  standard first-order tools.
  
  In previous work \cite{B9,C27,JSL} we have picked up and extended
  the embedding of multimodal logics in simple type theory as proposed
  by Brown~\cite{brown05:_encod_hybrid_logic_in_higher_order_logic}.
  The starting point is a characterization of multimodal logic
  formulas as particular $\lambda$-terms in simple type theory.
  A characteristic of the encoding is that the definiens of the
  $\mball{r}$ operator $\lambda$-abstracts over the accessibility
  relation~$r$. We have proved that this encoding is sound and
  complete \cite{C27,JSL} and we have illustrated that this encoding
  supports the formulation of meta properties of encoded multimodal
  logics such as the correspondence between certain axioms and
  properties of the accessibility relation \cite{B9}.  Some of these
  meta properties can even be effectively automated within our
  higher-order theorem prover LEO-II~\cite{C26}.
  
  In this paper we extend our previous work to quantified multimodal
  logics. Multimodal logics with quantification for propositional
  variables have been studied by others before, including Kripke
  \cite{Kripke59}, Bull \cite{Bull69}, Fine \cite{Fine70,Fine79},
  Kaplan \cite{Kaplan70}, and Kremer \cite{Kremer97}. Also first-order
  modal logics \cite{FM98,hughes96:_new_introd_to_modal_logic} have been studied in numerous
  publications.  We are interested here in multimodal logics with
  quantification over both propositional and first-order variables, a
  combination investigated, for example, by Fitting
  \cite{Fitting02}.  In contrast to Fitting we here pursue the
  translational approach and study the embedding of quantified
  multimodal logic in simple type theory. This approach has several
  advantages:
\begin{itemize}
\item The syntax and semantics of simple type theory is well
  understood \cite{Andrews72b,Andrews72a,BBK04,Henkin50}.
  Studying (quantified) multimodal logics as fragments of simple type
  theory can thus help to better understand semantical issues.
\item For simple type theory, various automated proof tools are available, including
  Isabelle/HOL \cite{NPW02}, HOL \cite{GordonMelham93},
  LEO-II \cite{C26}, and TPS \cite{DBLP:journals/japll/AndrewsB06}.
  Employing the transformation presented in this paper, these systems
  become immediately applicable to quantified multimodal logics
  or fragments of them.
\item Even meta properties of quantified modal logics can be
  formalized and mechanically analyzed within these provers. 
\item The systematic study of embeddings of multimodal logics in
  simple type theory can identify fragments of simple type theory that
  have interesting computational properties (such as the detection of
  the guarded fragment).  This can foster improvements to proof
  tactics in interactive proof assistants.
\end{itemize}

\noindent
Our paper is organized as follows. In Section\,\ref{sec2} we briefly
review simple type theory and adapt Fitting's \cite{Fitting02} notion
of quantified multimodal logics. In Section\,\ref{sec3} we extend our
previous work \cite{C27,B9,JSL} and present an embedding of quantified
multimodal logic in simple type theory. This embedding is shown sound
and complete in Section\,\ref{sec4}. In Section\,\ref{sec5} we present
some simple experiments with the automated theorem provers LEO-II, TPS, and
IsabelleP and the model finder IsabelleM. These experiments exploit
the new TPTP THF infrastructure \cite{BRS08}.

\section{Preliminaries} \label{sec2}
\subsection{Simple Type Theory}
Classical higher-order logic or \emph{simple type theory} $\stt$
\cite{Andrews2002a,Church40} is built on top of the simply
typed $\lambda$-calculus. The set~$\mathcal{T}$ of simple types is
usually freely generated from a set of basic types $\{o, \indtype\}$
(where $o$ is the type of Booleans and $\indtype$ is the type
of individuals) using the function type constructor
$\typearrow$. Instead of $\{o, \indtype\}$ we here consider a set of base
types $\{o, \indtype, \worldtype\}$, providing an additional base type
$\worldtype$ (the type of possible worlds).

The simple type theory language $\stt$ is defined by
\edcorr
{($\alpha$, $\beta$, $o\in\mathcal{T}$):}
{($\alpha$, $\beta\in\mathcal{T}$):}
\begin{eqnarray*} 
s,t & ::= & p_\alpha  \mid X_\alpha  \mid (\lam{X_\alpha} s_\beta)_{\alpha\typearrow\beta} \mid (s_{\alpha\typearrow\beta}\, t_\alpha)_\beta \mid (\neg_{o\typearrow o}\, s_o)_o \mid \\ & & (s_o \vee_{o\typearrow o \typearrow o} t_o)_o \mid 
(s_\alpha =_{\alpha\typearrow\alpha\typearrow o} t_\alpha)_o
\mid (\Pi_{(\alpha\typearrow o)\typearrow o}\, s_{\alpha\typearrow o})_o 
\end{eqnarray*}
$p_\alpha$ denotes typed constants and $X_\alpha$ typed variables
(distinct from $p_\alpha$).  Complex typed terms are constructed via
abstraction and application.  Our logical connectives of choice are
$\neg_{o\typearrow o}$, $\lor_{o\typearrow o\typearrow o}$, 
$=_{\alpha\typearrow\alpha\typearrow o}$ and
$\Pi_{(\alpha\typearrow o)\typearrow o}$ (for each type $\alpha$).
From these connectives, other logical connectives can be defined in
the usual way. We often use binder notation $\all{X_\alpha} s$ for
$\Pi_{(\alpha\typearrow o)\typearrow o}(\lam{X_\alpha} s_o)$.  We
denote \emph{substitution} of a term $A_\alpha$ for a variable $X_\alpha$ in
a term $B_\beta$ by $[A/X]B$.  Since we consider $\alpha$-conversion
implicitly, we assume the bound variables of $B$ avoid variable
capture.  Two common relations on terms are given by $\beta$-reduction
and $\eta$-reduction.  A $\beta$-redex has the form $(\lam{X}s)t$ and $\beta$-reduces
to $[t/X]s$.  An $\eta$-redex has the form $(\lam{X}s X)$ where variable $X$ is not
free in $s$; it $\eta$-reduces to $s$.  We write $s\eqb t$ to mean $s$
can be converted to $t$ by a series of $\beta$-reductions and
expansions.  Similarly, $s\eqbe t$ means $s$ can be converted to $t$
using both $\beta$ and $\eta$.
For each $s\in L$ there is a 
unique \emph{$\beta$-normal form} 
and a unique \emph{$\beta\eta$-normal form}. 
 
The semantics of $\stt$ is well understood and thoroughly documented
in the literature \cite{Andrews72b,Andrews72a,BBK04,Henkin50}; our
summary below is adapted from Andrews~\cite{sep-type-theory-church}.

A \emph{frame} is a collection $\{D_\alpha\}_{\alpha\in\mathcal{T}}$
of nonempty sets $D_\alpha$, such that $D_o = \{T, F\}$
(for truth and falsehood).  The
$D_{\alpha\typearrow\beta}$ are collections of functions mapping
$D_\alpha$ into $D_\beta$. The members of $D_\indtype$ are called
\emph{individuals}. An \emph{interpretation} is a tuple $\langle
\{D_\alpha\}_{\alpha\in\mathcal{T}}, I \rangle$ where function $I$
maps each typed constant $c_\alpha$ to an appropriate element of
$D_\alpha$, which is called the \emph{denotation} of $c_\alpha$ (the
logical symbols $\neg$, $\vee$, $\Pi^\alpha$, and $=_{\alpha\typearrow \alpha \typearrow o}$ are always given the standard
denotations). A \emph{variable assignment} $\phi$ maps variables
$X_\alpha$ to elements in $D_\alpha$.  An interpretation $\langle
\{D_\alpha\}_{\alpha\in{\mathcal{T}}}, I \rangle$ is a \emph{Henkin
  model} (equivalently, a \emph{general model}) if and only if there
is a binary function $\mathcal{V}$ such that $\mathcal{V}_\phi\,
s_\alpha \in D_\alpha$ for each variable assignment $\phi$ and term
$s_\alpha\in L$, and the following conditions are satisfied for all
$\phi$ and all $s,t\in L$: (a) $\mathcal{V}_\phi X_\alpha = \phi
X_\alpha$, (b) $\mathcal{V}_\phi\, p_\alpha = I p_\alpha$, (c)
$\mathcal{V}_\phi (s_{\alpha\typearrow\beta}\, t_\alpha) =
(\mathcal{V}_\phi\, s_{\alpha\typearrow\beta}) (\mathcal{V}_\phi
t_\alpha$), and (d) $\mathcal{V}_\phi (\lam{X_\alpha} s_\beta)$ is
that function from $D_\alpha$ into $D_\beta$ whose value for each
argument $z\in D_\alpha$ is $\mathcal{V}_{[z/X_\alpha]\phi} s_\beta$,
where ${[z/X_\alpha]\phi}$ is that variable assignment such that
$({[z/X_\alpha]\phi})X_\alpha = z$ and $({[z/X_\alpha]\phi}) Y_\beta =
\phi Y_\beta$ if $Y_\beta \not= X_\alpha$. (Since $I\/ \neg$, $I\/
\vee$, $I\/ \Pi$, and $I\/ {=}$ always denote the standard truth functions, we
have $\mathcal{V}_\phi\, (\neg s) = T$ if and only if $\mathcal{V}_\phi\, s = F$,
$\mathcal{V}_\phi\, (s \vee t) = T$ if and only if $\mathcal{V}_\phi\, s = T$ or
$\mathcal{V}_\phi\, t = T$, $\mathcal{V}_\phi\, (\all{X_\alpha}
s_o) = \mathcal{V}_\phi\, (\Pi^\alpha(\lam{X_\alpha} s_o)) = T$ if and only if
for all $z\in D_\alpha$ we have $\mathcal{V}_{[z/X_\alpha]\phi}\, s_o
= T$, and $\mathcal{V}_\phi\, (s = t) = T$ if and only if $\mathcal{V}_\phi\, s = \mathcal{V}_\phi\, t$. Moreover, we have $\mathcal{V}_\phi\, s = \mathcal{V}_\phi\, t$
whenever $s\eqbe t$.) It is easy to verify that Henkin models obey the 
\edcorr
{``Denotatpflicht'',}
{rule that everything denotes,}
that is, each term $t_\alpha$
always has a denotation $\mathcal{V}_\phi\,t_\alpha \in D_\alpha$.
  If an interpretation $\langle \{D_\alpha\}_{\alpha\in\mathcal{T}}, I
  \rangle$ is a Henkin model, then the function $\mathcal{V}_\phi$ is
  uniquely determined.

  We say that formula $A\in L$ is \emph{valid in a model} $\langle
  \{D_\alpha\}_{\alpha\in\mathcal{T}}, I \rangle$ if and only if
  $\mathcal{V}_\phi A = T$ for every variable assignment $\phi$. A
  model for a set of formulas $H$ is a model in which each formula of
  $H$ is valid.  A formula $A$ is Henkin-valid if and only if $A$ is
  valid in every Henkin model.
 We write $\models^{\tt} A$ if $A$ is Henkin-valid. 

\vfill\subsection{Quantified Multimodal Logic} 
 
First-order quantification can be constant domain or varying domain.  Below we only consider the constant domain case: every possible world has the same domain. We adapt the presentation of syntax and
 semantics of quantified modal logic from Fitting \cite{Fitting02}.
In contrast to Fitting we are not interested in \textbf{S5} structures but in the more general case
of \textbf{K}.

Let $\IV$ be a set of first-order (individual) variables, $\PV$ a
set of propositional variables, and $\SYM$ a set of predicate
symbols of any arity. Like Fitting, we keep our definitions simple by not having function or constant symbols.
  While Fitting \cite{Fitting02} studies quantified monomodal logic, we
  are interested in quantified multimodal logic. Hence, we introduce
  multiple $\mball{r}$ operators for symbols $r$ from an index set
  $S$. The grammar for our quantified multimodal logic $\QML$ is thus
\begin{eqnarray*} 
  s,t & ::= & P \mid k(X^1,\ldots,X^n) \mid \mnot s \mid s \mor t \mid \all{X} s  \mid \all{P} s  \mid \mball{r} s 
\end{eqnarray*}
where $P\in \PV$, $k\in\SYM$, and $X,X^i\in\IV$.  

Further connectives, quantifiers, and modal operators can be defined
as usual.  We also obey the usual definitions of free variable
occurrences and substitutions.

Fitting introduces three different notions of semantics:
\textbf{QS5$\pi^-$}, \textbf{QS5$\pi$}, and \textbf{QS5$\pi^+$}.  We
study related notions $\QKPIm$, $\QKPI$, and $\QKPIp$ for a modal
context \textbf{K}, and we support multiple modalities.

A \textit{$\QKPIm$ model} is a structure $M=(W,(R_r)_{r\in S},D,P,(I_w)_{w\in W})$ such that
$(W,(R_r)_{r\in S})$ is a multimodal frame (that is, $W$ is the set of possible
worlds and the $R_r$ are accessibility relations between worlds in
$W$), $D$ is a non-empty set (the first-order domain), $P$ is a
non-empty collection of subsets of $W$ (the propositional domain), and
the $I_w$ are interpretation functions mapping each $n$-place relation symbol
$k\in\SYM$ to some $n$-place relation on $D$ in world $w$.

A \textit{variable assignment} $g=(g^{iv},g^{pv})$ is a pair of maps $g^{iv}: \IV \longrightarrow D$ and  $g^{pv}: \PV \longrightarrow P$,
where $g^{iv}$ maps each individual variable in $\IV$ to a an object in $D$ and $g^{pv}$ maps each
propositional variable in $\PV$ to a set of worlds in $P$.

Validity of a formula $s$ for a model $M=(W,(R_r)_{r\in S},D,P,I_w)$,
a world $w\in W$, and a variable assignment $g=(g^{iv},g^{pv})$ is denoted as $M,g,w\models s$ and defined as
follows, where $[a/Z]g$ denotes the assignment identical to $g$ except that 
\edcorr
{$g(Z) = a$:}
{$([a/Z]g)(Z) = a$:\vfill\pagebreak}
\begin{eqnarray*}
M,g,w \models k(X^1,\ldots,X^n) & \text{ if and only if } & \langle g^{iv}(X^1),\ldots,g^{iv}(X^n)\rangle\in I_w(k) \\
M,g,w \models P & \text{ if and only if } & w\in g^{pv}(P) \\
M,g,w \models \mnot p & \text{ if and only if } & M,g,w \not\models p\\
M,g,w \models p \mor q & \text{ if and only if } & M,g,w \models p \text{ or } M,g,w \models q \\
M,g,w \models \mall{X}p & \text{ if and only if } & M,([d/X]g^{iv},g^{pv}),w \models p \text{ for all } d\in D \\
\edcorr
{M,g,w \models \mall{P}p & \text{ if and only if } & M,(g^{iv},[v/P]g^{pv}),w \models p \text{ for all } v\in P \\}
{M,g,w \models \mall{Q}p & \text{ if and only if } & M,(g^{iv},[v/Q]g^{pv}),w \models p \text{ for all } v\in P \\}
M,g,w \models \mball{r}p & \text{ if and only if } & M,g,v \models p \text{ for all } v\in W \\
& & \text{ with } \langle w,v\rangle\in R_r
\end{eqnarray*}

A $\QKPIm$ model $M=(W,(R_r)_{r\in S},D,P,(I_w)_{w\in W})$ is a
\textit{$\QKPI$ model} if for every variable assignment $g$ and every formula
$s\in\QML$, the set of worlds $\{w\in W \mid M,g,w \models s\}$ is a member of
$P$. 

A $\QKPI$ model $M=(W,(R_r)_{r\in S},D,P,(I_w)_{w\in W})$ is a
\textit{$\QKPIp$ model} if every world $w\in W$ is member of an atom
in $P$. The \textit{atoms} of $P$ are minimal non-empty elements of $P$:
no proper subsets of an atom are also elements of $P$.
 
A $\QML$ formula $s$ is \textit{valid in model $M$ for world
  $w$} if $M,g,w \models s$ for all variable assignments $g$. A
formula $s$ is \textit{valid in model $M$} if $M,g,w \models s$
for all $g$ and $w$. Formula $s$ is \textit{$\QKPI$-valid} if $s$ is valid in all $\QKPI$ models, when we
write $\models^{\QKPI} s$; we define $\QKPIm$-valid and $\QKPIp$-valid analogously.

In the remainder we mainly focus on $\QKPI$ models. 
\edcorr
{These}
{These models}
naturally correspond to Henkin models, as we shall see in Sect.\ref{sec4}.

\vfill\pagebreak

\section{Embedding Quantified Multimodal Logic in $\stt$} \label{sec3} 
The idea of the encoding is simple. We
choose type $\indtype$ to denote the (non-empty) set of
individuals and we reserve a second base type $\worldtype$ to denote
the (non-empty) set of possible worlds. 
The type $o$ denotes the set of truth values.  Certain formulas of
type $\worldtype\typearrow o$ then correspond to multimodal logic
expressions. The multimodal connectives $\mnot$, $\mor$, and
$\mball{}$, become $\lambda$-terms of types
${(\worldtype\typearrow o)\typearrow(\worldtype\typearrow o)}$,
${(\worldtype\typearrow o)\typearrow(\worldtype\typearrow o)
  \typearrow (\worldtype\typearrow o)}$, and
${(\worldtype\typearrow\worldtype\typearrow
  o)\typearrow(\worldtype\typearrow o) \typearrow (\worldtype
  \typearrow o)}$ respectively.

Quantification is handled as usual in higher-order logic by modeling
$\all{X} p$ as $\Pi(\lam{X}p)$ for a suitably chosen connective
$\Pi$, as we remarked in Section\,\ref{sec2}.  Here we are interested in defining two
particular modal $\modal\Pi$-connectives: $\modal\Pi^\indtype$, for
quantification over individual variables, and
$\modal\Pi^{\worldtype\typearrow o}$, for quantification over modal
propositional variables that depend on worlds,
of types $(\indtype\typearrow(\worldtype \typearrow o))\typearrow(\worldtype \typearrow
o)$ and $((\worldtype\typearrow o)\typearrow(\worldtype \typearrow
o))\typearrow(\worldtype \typearrow o)$, respectively.

In previous work \cite{B9} we have discussed first-order and
higher-order modal logic, including a means of
explicitly excluding terms of certain types. The idea was that no proper
subterm of $t_{\worldtype\typearrow o}$ should introduce a dependency on
worlds. Here we skip this restriction.  This leads to a simpler
definition of a quantified multimodal language $\QMLSTT$ below, and it
does not affect our soundness and completeness results.



\begin{definition}[Modal operators]\label{operators}\\
The modal operators $\mnot,\mor,\modal\Box,\modal\Pi^\indtype$, and $\modal\Pi^{\worldtype \typearrow o}$ are defined as follows:
\begin{align*}
\mnot_{(\worldtype\typearrow o)\typearrow(\worldtype\typearrow o)} & = \lam{\phi_{\worldtype\typearrow o}}\lam{W_\worldtype}\neginpit{\phi\,W}
\\
\mor_{(\worldtype\typearrow o)\typearrow(\worldtype\typearrow o)\typearrow(\worldtype\typearrow o)} & = \lam{\phi_{\worldtype\typearrow o}} \lam{\psi_{\worldtype\typearrow o}} \lam{W_\worldtype} \phi\,W \vee \psi\,W
\\
\mball{}_{(\worldtype\typearrow\worldtype\typearrow o)\typearrow(\worldtype\typearrow o)\typearrow(\worldtype\typearrow o)} & =  
\lam{R_{\worldtype\typearrow\worldtype\typearrow o}} \lam{\phi_{\worldtype\typearrow o}} \lam{W_{\worldtype}} \all{V_{\worldtype}} \neginpit{R\,W\,V} \vee \phi\,V 
\\
\modal\Pi^\indtype_{(\indtype\typearrow(\worldtype \typearrow o))\typearrow(\worldtype \typearrow o)} & = \lam{\phi_{\indtype\typearrow(\worldtype \typearrow
    o)}} \lam{W_\worldtype} \all{X_\indtype} \phi\,X\,W \\
\modal\Pi^{\worldtype \typearrow o}_{((\worldtype \typearrow o)\typearrow(\worldtype \typearrow o))\typearrow(\worldtype \typearrow o)} & = \lam{\phi_{(\worldtype \typearrow o)\typearrow(\worldtype \typearrow
    o)}} \lam{W_\worldtype} \all{P_{\worldtype\typearrow o}} \phi\,P\,W
\end{align*}
Further operators can be introduced, for example,
\begin{align*}
\mtrue_{(\worldtype\typearrow o)\typearrow(\worldtype\typearrow o)} & = \all{P_{\worldtype\typearrow o}} P \mor \mnot P\\
\mfalse_{(\worldtype\typearrow o)\typearrow(\worldtype\typearrow o)} & = \mnot \mtrue \\
\mand_{(\worldtype\typearrow o)\typearrow(\worldtype\typearrow o)\typearrow(\worldtype\typearrow o)} & = \lam{\phi_{\worldtype\typearrow o}} \lam{\psi_{\worldtype\typearrow o}} \mnot (\mnot \phi \mor \mnot \psi)\\
\mimpl_{(\worldtype\typearrow o)\typearrow(\worldtype\typearrow o)\typearrow(\worldtype\typearrow o)}
&=\lam{\phi_{\worldtype\typearrow o}} \lam{\psi_{\worldtype\typearrow o}} \mnot \phi \mor \psi
\\
\mdexi{}_{(\worldtype\typearrow\worldtype\typearrow o)\typearrow(\worldtype\typearrow o)\typearrow(\worldtype\typearrow o)} 
&=\lam{R_{\worldtype\typearrow\worldtype\typearrow o}}\lam{\phi_{\worldtype\typearrow o}} 
\mnotinpit{\modal\Box\,R\,(\mnot \phi)}
\\\modal\Sigma^\indtype_{(\indtype\typearrow(\worldtype \typearrow o))\typearrow(\worldtype \typearrow o)}
&=\lam{\phi_{\indtype\typearrow(\worldtype\typearrow o)}}\mnotinpit{\modal\Pi^\indtype (\lam{X_\indtype} \mnotinpit{\phi\,X})} 
\\\modal\Sigma^{\worldtype \typearrow o}_{((\worldtype \typearrow o)\typearrow(\worldtype \typearrow o))\typearrow(\worldtype \typearrow o)}
&=\lam{\phi_{(\worldtype \typearrow o)\typearrow(\worldtype\typearrow o)}}\mnotinpit
{\modal\Pi^{\worldtype \typearrow o}(\lam{P_{\worldtype\typearrow o}}\mnotinpit{\phi\,P})}
\end{align*}
\end{definition}

\vfill\vspace{\topsep}
\noindent
We could also introduce further modal operators, such as the
difference modality $D$, the global modality $E$, nominals with $!$,
or the $@$ operator (consider the recent work of Kaminski and Smolka
\cite{KaminskiSmolka08converse} in the propositional hybrid logic
context; they also adopt a higher-order perspective):
\begin{align*}
D_{(\worldtype\typearrow o)\typearrow(\worldtype\typearrow o)} & = \lam{\phi_{\worldtype\typearrow o}} \lam{W_\worldtype} \exi{V_\worldtype} W \not= V \wedge \phi V 
\\
E_{(\worldtype\typearrow o)\typearrow(\worldtype\typearrow o)} & = \lam{\phi_{\worldtype\typearrow o}} \phi \mor D\, \phi 
\\
!_{(\worldtype\typearrow o)\typearrow(\worldtype\typearrow o)} & = \lam{\phi_{\worldtype\typearrow o}} E\, (\phi \mand \mnotinpit{D\,\phi}) 
\\
@_{\worldtype\typearrow(\worldtype\typearrow o)\typearrow(\worldtype\typearrow o)} & = \lam{W_\worldtype} \lam{\phi_{\worldtype\typearrow o}} \phi\, W 
\end{align*}
This illustrates the potential of our embedding for encoding quantified
hybrid logic, an issue that we might explore in future work.

For defining $\QMLSTT$-propositions we fix a set $\IVSTT$ of individual variables of type $\indtype$, a set $\PVSTT$
of propositional variables of type $\worldtype\typearrow o$, and a set $\SYMSTT$ of $k$-ary (curried) predicate constants of types ${\text{${\underbrace{\indtype\typearrow\ldots \typearrow\indtype}_{n}\typearrow(\worldtype\typearrow o)}$}}$. 
The latter types will be abbreviated as  ${\indtype^n\typearrow(\worldtype\typearrow o)}$ in the remainder. Moreover, we fix a set $\SSTT$ of accessibility relation constants of type $\worldtype\typearrow \worldtype \typearrow o$. 


\begin{definition}[$\QMLSTT$-propositions]\label{qmlstt}\sloppy
$\QMLSTT$-propositions are defined as the smallest set of simply typed $\lambda$-terms for  which the following hold:
\begin{itemize}
\item Each variable $P_{\worldtype\typearrow o}\in\PVSTT$ is an atomic $\QMLSTT$-proposition, and if $X_\indtype^j\in\IVSTT$ (for $j=1$, \ldots, $n$) and $k_{\indtype^n\typearrow(\worldtype\typearrow o)}\in\SYMSTT$, then the term $(k\,X^1 \, \ldots \, X^n)_{\worldtype\typearrow o}$ is an atomic $\QMLSTT$-proposition.
\item If $\phi$ and $\psi$ are $\QMLSTT$-propositions, then so are $\mnot\,\phi$ and 
  $\phi\mor\psi$.
\item If $r_{\worldtype\typearrow\worldtype\typearrow o}\in\SSTT$ is an accessibility relation constant and if $\phi$ is an $\QMLSTT$-proposition, then $\modal\Box\,r\,\phi$  is a  $\QMLSTT$-proposition. 
\item If $X_\indtype\in\IVSTT$ is an individual variable and $\phi$ is a $\QMLSTT$-proposition then  $\modal\Pi^\indtype(\lam{X_\indtype} \phi)$ is a  $\QMLSTT$-proposition.
\item If $P_{\worldtype\typearrow o}\in\PVSTT$  is a propositional variable and $\phi$ is a $\QMLSTT$-proposition then  $\modal\Pi^{\worldtype\typearrow o}(\lam{P_{\worldtype\typearrow o}} \phi)$ is a  $\QMLSTT$-proposition. 
\end{itemize}
We write
  $\mball{r}\phi$, $\all{X_\indtype}\phi$, and  $\all{P_{\worldtype\typearrow o}}\phi$ for $\modal\Box\,r\,\phi$, $\modal\Pi^\indtype(\lam{X_\indtype} \phi)$, and $\modal\Pi^{\worldtype\typearrow o}(\lam{P_{\worldtype\typearrow o}} \phi)$, respectively.
\end{definition}

\vfill\vspace{\topsep}\noindent
Because the defining equations in Definition\,\ref{operators} 
are themselves formulas
in simple type theory, we can express proof problems
in a higher-order theorem prover 
elegantly in the syntax of quantified multimodal logic.  Using
rewriting or definition expanding, we can reduce these
representations to corresponding statements containing only
the basic connectives $\mnot$, $\mor$, $=$, $\Pi^\indtype$, and $\Pi^{\worldtype\typearrow o}$ of simple type theory.

\vspace{\topsep}\vspace{\topsep}
\begin{example}\label{ex1}
  The following $\QMLSTT$ proof problem expresses that in all accessible
  worlds there exists truth:
\[\mball{r} \mexi{P_{\worldtype\typearrow o}} P\]
The term rewrites into the following $\beta\eta$-normal term of type
$\worldtype \typearrow o$
\[\lam{W_\worldtype} \all{Y_\worldtype}\neginpit{r\,W\,Y}\vee (\neg \all{P_{\worldtype\typearrow o}} \neg (P\, Y)) \]
\end{example}

\vfill\pagebreak

\noindent
Next, we define validity of $\QMLSTT$ propositions
$\phi_{\worldtype\typearrow o}$ in the obvious way: a $\QML$-proposition
$\phi_{\worldtype\typearrow o}$ is valid if and only if for all possible worlds
$w_\worldtype$ we have 
\edcorr
{$w_\worldtype\worldtype\in \phi_{\worldtype\typearrow o}$,}
{$w_\worldtype\in \phi_{\worldtype\typearrow o}$,}
that is, if and only if
$\phi_{\worldtype\typearrow o}\,w_\worldtype$ holds.

\begin{definition}[Validity]\label{validity}\\
  Validity is modeled as an abbreviation for the following simply typed
  $\lambda$-term:
\begin{eqnarray*}
\text{valid} & = & \lam{\phi_{\worldtype\typearrow o}} \all{W_{\worldtype}} \phi\,W \\
\end{eqnarray*}
\end{definition}

\begin{example} \label{ex2} \sloppy
 We analyze whether the proposition $\mball{r} \mexi{P_{\worldtype\typearrow o}} P$ is valid or not. For this, we
 formalize the following proof problem
\[\text{valid}\,\, (\mball{r} \mexi{P_{\worldtype\typearrow o}} P) \]
Expanding this term leads to
\[\all{W_\worldtype} \all{Y_\worldtype} \neginpit{r\,W\,Y} \vee (\neg \all{X_{\worldtype\typearrow o}} \neg (X\, Y)) \] 
It is easy to check that this term is valid in Henkin semantics: put $X = \lam{Y_\worldtype} \mtrue$. 
\end{example}

\vspace{\topsep}\noindent
An obvious question is whether the notion of quantified 
multimodal logics we obtain via this embedding indeed exhibits the desired
properties. In the next section, we prove soundness and
completeness for a mapping of $\QML$-propositions to $\QMLSTT$-propositions.


\vfill\pagebreak

\section{Soundness and Completeness of the Embedding} \label{sec4}
In our soundness proof, we exploit the following mapping of $\QKPI$ models into Henkin models.
We assume that the $\QML$ logic $L$ under consideration is constructed as outlined in Section\,\ref{sec2}
from a set  of individual variables $\IV$, a set of propositional variables
$\PV$, and a set of predicate symbols $\SYM$. Let 
  $\mball{r^1}$, \ldots, $\mball{r^n}$ for $r^i\in S$ be the box operators of $L$.

\begin{definition}[$\QMLSTT$ logic $L^\STT$ for $\QML$ logic $L$]\label{def1}\\
Given an $\QML$ logic $L$, define a mapping $\dot{\_}$ as follows:
\begin{align*}
\dot{X} & = X_\indtype \text{ for every } X \in \IV \\
\dot{P} & = P_{\worldtype\typearrow o}  \text{ for every } P \in \PV  \\
\dot{k} & = k_{\indtype^n\typearrow(\worldtype\typearrow o)} \text{ for n-ary } k \in \SYM \\
\dot{r} & = r_{\worldtype\typearrow \worldtype\typearrow o} \text{ for every } r \in S 
\end{align*}
The $\QMLSTT$ logic $L^\STT$ is obtained from $L$ by applying Def.\,\ref{qmlstt}
with $\IVSTT = \{\dot{X} \mid X \in \IV\}$, $\PVSTT
  = \{\dot{P}\mid P \in \PV\}$, $\SYMSTT = \{\dot{k} \mid k\in\SYM\}$, and $\SSTT = \{\dot{r} \mid r \in S\}$.
Our construction obviously induces a one-to-one correspondence $\dot{\_}$ between languages $L$ and $L^\STT$.


Moreover, let $g = (g^{iv}:\IV\longrightarrow D,\, g^{pv}:\PV\longrightarrow
P)$ be a variable assignment for $L$.  We define the corresponding variable assignment
$$\dot{g}=(\dot{g}^{iv}:\IVSTT\longrightarrow
D = D_\indtype,\;\dot{g}^{pv}:\PVSTT\longrightarrow P = D_{\worldtype\typearrow o})$$ for $L^\STT$ so that $\dot{g}(X_\indtype) =
\dot{g}(\dot{X}) = g(X)$ and $\dot{g}(P_{\worldtype\typearrow o}) =
\dot{g}(\dot{P}) = g(P)$ for all $X_\indtype\in\IVSTT$ and
$P_{\worldtype\typearrow o}\in\PVSTT$.

Finally, a variable assignment $\dot{g}$ is lifted to an assignment for variables $Z_\alpha$ of arbitrary type by choosing $\dot{g}(Z_\alpha) = d\in D_\alpha$ arbitrarily, if $\alpha\not= \indtype,{\worldtype\typearrow o}$.
\end{definition}

\vspace{\topsep}\noindent
We assume below that $L$, $L^\STT$, $g$ and $\dot{g}$ are defined as above.

\begin{definition}[Henkin model $H^Q$ for $\QKPI$ model $Q$] \label{def2} \\
Given a $\QKPI$ model $Q=(W,(R_r)_{r\in
    S},D,P,(I_w)_{w\in W})$ for $L$, a Henkin
  model $H^Q=\langle
\{D_\alpha\}_{\alpha\in{\mathcal{T}}}, I \rangle$ for  $L^\STT$ is constructed as follows. We choose 
\begin{itemize}
\item the set $D_\worldtype$ as the set of  possible worlds $W$,
\item the set $D_\indtype$ as the set of individuals $D$ (cf.\ definition of $\dot{g}^{iv}$),
\item the set $D_{\worldtype\typearrow o}$ as the set of sets of possible worlds $P$ (cf.\ definition of $\dot{g}^{pv}$),\footnote{To keep things simple, we identify sets with their characteristic functions.}
\item the set $D_{\worldtype\typearrow \worldtype \typearrow o}$ as the set of relations $(R_r)_{r\in S}$,
\item and all other sets $D_{\alpha\typearrow\beta}$ as (not necessarily full)  sets of functions from $D_\alpha$ to $D_\beta$; for all sets $D_{\alpha\typearrow\beta}$ the 
\edcorr
{``Denotatpflicht''}
{rule that everything denotes}
must be obeyed, in particular, we require that the sets $D_{\indtype^n\typearrow(\worldtype\typearrow o)}$ contain the elements $I k_{\indtype^n\typearrow(\worldtype\typearrow o)}$ as characterized below.
\end{itemize}
The interpretation $I$ is as follows: 
\begin{itemize} \sloppy
\item Let $k_{\indtype^n\typearrow(\worldtype\typearrow o)} =\dot{k}$ for
  $k\in\SYM$ and let $X^i_\indtype = \dot{X^i}$ for $X^i\in\IV$.  We choose
  $I k_{\indtype^n\typearrow(\worldtype\typearrow o)} \in
  D_{\indtype^n\typearrow(\worldtype\typearrow o)}$ such that
  $$(I\,k)(\dot{g}(X^1_\indtype),\ldots,\dot{g}(X^n_\indtype),w) = T$$ for all
  worlds $w\in D_\worldtype$ such that $Q,g,w\models k(X^1,\ldots,X^n)$; that
  is, if $\langle g(X^1),\ldots,g(X^n)\rangle\in I_w(k)$.   Otherwise $(I\,k)
  (\dot{g}(X^1_\indtype),\ldots,\dot{g}(X^n_\indtype),w) = F$.
\item Let $r_{\worldtype\typearrow\worldtype\typearrow o} = \dot{r}$ for $r\in S$. 
  We choose $I r_{\worldtype\typearrow\worldtype\typearrow o}\in
  D_{\worldtype\typearrow\worldtype\typearrow o}$ such that $(I
  r_{\worldtype\typearrow\worldtype\typearrow o})(w,w') = T$ if
  $\langle w,w'\rangle\in R_{r}$ in  $Q$ and $(I
  r_{\worldtype\typearrow\worldtype\typearrow o})(w,w') = F$ otherwise.
\end{itemize}
\end{definition}

 It is not hard to verify  that $H^Q=\langle \{D_\alpha\}_{\alpha\in\mathcal{T}},
  I \rangle$ is a Henkin model. 

\begin{lemma} \label{lemma1}\\ 
  Let $Q=(W,(R_r)_{r\in S},D,P,(I_w)_{w\in W})$ be a $\QKPI$ model and
  let $H^Q=\langle \{D_\alpha\}_{\alpha\in\mathcal{T}}, I \rangle$ be
  a Henkin model for $Q$. 
Furthermore, let $s_{\worldtype\typearrow o} = \dot{s}$ for 
\edcorr
{$s\in\L$.} 
{$s\in L$.}
\\ 
Then for all worlds $w\in W$ and variable assignments $g$ we have $Q,g,w
  \models s$ in $Q$ if and only if $\mathcal{V}_{[w/W_\worldtype]\dot{g}}\,
  (s_{\worldtype\typearrow o} W_\worldtype) = T$ in $H^Q$.
\end{lemma}

\textbf{Proof:} \sloppy The proof is by induction on the structure of $s\in
  L$. 

  Let $s=P$ for $P\in\PV$. By
  construction of Henkin model $H^Q$ and by definition of $\dot{g}$, we
  have for $P_{\worldtype\typearrow o}=\dot{P}$ that
  $\mathcal{V}_{[w/W_\worldtype]\dot{g}}\, (P_{\worldtype\typearrow o}\, W_\worldtype) =
  \dot{g}(P_{\worldtype\typearrow o})(w) = T$ if and only if $Q,g,w \models
  P$, that is, $w\in g(P)$.

  Let $s=k(X^1,\ldots,X^n)$ for $k\in\SYM$ and $X^i\in\IV$. By
  construction of Henkin model $H^Q$ and by definition of $\dot{g}$, we have for
  $\dot{k}(\dot{X}^1,\ldots,\dot{X}^n) =
  (k_{\indtype^n\typearrow(\worldtype\typearrow o)}\, X^1_\indtype\, \ldots\,
  X^n_\indtype)$ that 
  $$\mathcal{V}_{[w/W_\worldtype]\dot{g}}\,
  ((k_{\indtype^n\typearrow(\worldtype\typearrow o)} X^1_\indtype \ldots
  X^n_\indtype)\,W_\worldtype) = (I\,k)(\dot{g}(X^1_\indtype),\ldots,\dot{g}(X^n_\indtype),w)
  = T$$ if and only if $Q,g,w \models k(X^1,\ldots,X^n)$, that is, $\langle g(X^1), \ldots, g(X^n) \rangle \in I_w(k)$.

  Let $s=\mnot t$ for $t\in L$. We have $Q,g,w \models \neg s$ if and only
  $Q,g,w \not\models s$, which is equivalent by induction to
  $\mathcal{V}_{[w/W_\worldtype]\dot{g}}\, (t_{\worldtype\typearrow o}\, W_\worldtype) =
  F$ and hence to $\mathcal{V}_{[w/W_\worldtype]\dot{g}}\, \neg
  (t_{\worldtype\typearrow o}\,W_\worldtype) =_{\beta\eta}
  \mathcal{V}_{[w/W_\worldtype]\dot{g}}\, ((\mnot\, t_{\worldtype\typearrow
    o})\, W_\worldtype) = T$.

  Let $s=(t\mor l)$ for $t,l\in L$. We have $Q,g,w \models (t\mor l)$ if
  and only if $Q,g,w \models t$ or $Q,g,w \models l$.  The latter
  condition is equivalent by induction to
  $\mathcal{V}_{[w/W_\worldtype]\dot{g}}\, (t_{\worldtype\typearrow o}\, \,
  W_\worldtype) = T$ or $\mathcal{V}_{[w/W_\worldtype]\dot{g}}\,
  (l_{\worldtype\typearrow o}\, \, W_\worldtype) = T$ and therefore to
  $\mathcal{V}_{[w/W_\worldtype]\dot{g}}\, (t_{\worldtype\typearrow o}\, \,
  W_\worldtype) \vee (l_{\worldtype\typearrow o} \, W_\worldtype) =_{\beta\eta}
  \mathcal{V}_{[w/W_\worldtype]\dot{g}}\, ( t_{\worldtype\typearrow o}\, \mor
  l_{\worldtype\typearrow o}\, \, W_\worldtype) = T$.

  Let $s=\mball{r} t$ for $t\in L$. We have $Q,g,w \models \mball{r}
  t$ if and only if for all $u$ with $\langle w,u\rangle \in R_r$ we
  have $Q,g,u\models t$. The latter condition is equivalent by
  induction to this one: for all $u$ with $\langle w,u\rangle \in R_r$
  we have $\mathcal{V}_{[u/V_\worldtype]\dot{g}}\, (t_{\worldtype\typearrow
    o}\, V_\worldtype) = T$. That is equivalent to
  $$\mathcal{V}_{[u/V_\worldtype][w/W_\worldtype]\dot{g}}\, (\neg
  (r_{\worldtype\typearrow\worldtype\typearrow o}\, W_\worldtype\, V_\worldtype) \vee
  (t_{\worldtype\typearrow o}\, V_\worldtype)) = T$$ 
  and thus to
  $$\mathcal{V}_{[w/W_\worldtype]\dot{g}}\, (\all{Y_\worldtype} (\neg
  (r_{\worldtype\typearrow\worldtype\typearrow o}\, W_\worldtype\, Y_\worldtype) \vee
  (t_{\worldtype\typearrow o}\, Y_\worldtype))) =_{\beta\eta}
  \mathcal{V}_{[w/W_\worldtype]\dot{g}}\, ( \mball{r} t \, W_\worldtype) = T.$$

  Let $s=\mall{X} t$ for $t\in L$ and $X\in\IV$. We have $Q,g,w
  \models \mall{X} t$ if and only if $Q,[d/X]g,w \models
  t$ for all $d\in D$. The latter condition is equivalent by induction
  to 
  $\mathcal{V}_{[d/X_\indtype][w/W_\worldtype]\dot{g}}\, (t_{\worldtype\typearrow
    o}\, W_\worldtype) = T$ for all $d\in D_\indtype$.
  That condition is equivalent to  $\mathcal{V}_{[w/W_\worldtype]\dot{g}}\,
  (\Pi^\indtype_{(\indtype\typearrow o)\typearrow o} (\lam{X_\indtype} t_{\worldtype\typearrow
    o}\, W_\worldtype)) =_{\beta\eta}\, \mathcal{V}_{[w/W_\worldtype]\dot{g}}\,
((\lam{V_\worldtype}\, (\Pi^\indtype_{(\indtype\typearrow o)\typearrow o}\, (\lam{X_\indtype} t_{\worldtype\typearrow
  o}\, V_\worldtype)))\, W_\worldtype) = T$ and so by definition of $\modal\Pi^\indtype$ to $\mathcal{V}_{[w/W_\worldtype]\dot{g}}\,
  ((\modal\Pi^\indtype_{(\indtype\typearrow (\worldtype\typearrow o))\typearrow
    (\worldtype\typearrow o)}\, (\lam{X_\indtype} t_{\worldtype\typearrow o}))\, W_\worldtype) = 
  \mathcal{V}_{[w/W_\worldtype]\dot{g}}\,
  ((\mall{X_\indtype} t_{\worldtype\typearrow o})\, W_\worldtype) = T$.

  The case for $s=\mall{P} t$ where $t\in L$ and $P\in\PV$ is analogous to $s=\mall{X} t$.
$\Box$

We exploit this result to prove the soundness of our embedding.

\begin{theorem}[Soundness for $\QKPI$ semantics] \label{thm1} 
  Let $s\in L$ be a $\QML$ proposition and let $s_{\worldtype\typearrow o} = \dot{s}$ be the
  corresponding $\QMLSTT$ proposition. If $\models^{\stt}
  (\text{valid}\, s_{\worldtype\typearrow o})$  then $\models^{\QKPI} s$. 
\end{theorem}

\textbf{Proof:}
  By contraposition, assume $\not\models^{\QKPI} s$:
  that is, there is a $\QKPI$ model $Q=(W,(R_r)_{r\in
    S},D,P,(I_w)_{w\in W})$, a variable assignment $g$ and a world $w\in W$, such that 
  $Q,g,w\not\models s$.  By Lemma \ref{lemma1}, 
  we have $\mathcal{V}_{[w/W_\worldtype]\dot{g}}\, (s_{\worldtype\typearrow o} \,
  W_\worldtype) = F$ in a Henkin model $H^Q$ for $Q$.  Thus, $\mathcal{V}_{\dot{g}}\,
  (\all{W_\worldtype} (s_{\worldtype\typearrow o}\, W)) =_{\beta\eta} \mathcal{V}_{\dot{g}}\, (\text{valid}\, s_{\worldtype\typearrow o}) = F$. Hence, $\not\models^{\tt} (\text{valid}\, s_{\worldtype\typearrow o})$.
$\Box$

In order to prove completeness, we reverse our mapping from Henkin
models to $\QKPI$ models.

\begin{definition}[$\QML$ logic $L^\QML$ for $\QMLSTT$ logic $L$]
  The mapping $\bar{\_}$ is defined as the reverse map of $\dot{\_}$ from 
Def.\,\ref{def1}.

  The $\QML$ logic $L^\QML$ is
  obtained from $\QMLSTT$ logic $L$ by choosing $\IV=\{\bar{X}_\indtype \mid X_\indtype \in
  \IVSTT\}$, $\PV=\{\bar{P}_{\worldtype\typearrow o} \mid
  P_{\worldtype\typearrow o} \in \PVSTT\}$,
  $\SYM=\{\bar{k}_{\indtype^n\typearrow(\worldtype\typearrow o} \mid
  k_{\indtype^n\typearrow(\worldtype\typearrow o)} \in \SYMSTT\}$, and $S=\{\bar{r}_{\worldtype\typearrow\worldtype\typearrow o} \mid r_{\worldtype\typearrow\worldtype\typearrow o}\in \SSTT\}$.

 Moreover, let $g:  \IVSTT\cup\PVSTT \longrightarrow D\cup P$ be a variable assignment for $L$.
 The corresponding variable assignment  $\bar{g}: \IV\cup\PV \longrightarrow D\cup P$ for $L^\QML$ is defined as follows:
 $\bar{g}(X) = \bar{g}(\bar{X_\indtype})=g(X_\indtype)$ and  $\bar{g}(P) = \bar{g}(\bar{P_{\worldtype\typearrow o}})=g(P_{\worldtype\typearrow o})$ for all $X\in\IV$ and $P\in\PV$.

\end{definition}

We assume below that $L$, $L^\QML$, $g$ and $\bar{g}$ are defined as above.

\begin{definition}[$\QKPIm$ model $Q^H$ for Henkin model
  $H$] \label{def3} Given a Henkin model $H=\langle
  \{D_\alpha\}_{\alpha\in\mathcal{T}}, I \rangle$ for $\QMLSTT$ logic
  $L$,  we construct a $\QML$ model $Q^H=(W,(R_r)_{r\in S},D,P,(I_w)_{w\in W})$
  for $L^\QML$ by choosing $W=D_\worldtype$,
  $D=D_\indtype$, $P=D_{\worldtype\typearrow o}$\footnote{Again, we identify sets with their characteristic functions.}, and $(R_r)_{r\in S}=
  D_{\worldtype\typearrow \worldtype\typearrow o}$.  Let
  $k=\bar{k}_{\indtype^n\typearrow(\worldtype\typearrow o)}$ and let $X^i =
  \bar{X}^i_\indtype$.  We choose $I_w(k)$ such that $\langle \bar{g}(X^1),
  \ldots, \bar{g}(X^n)\rangle\in I_w(k)$ if and only if
  $$(I\,k)(g(X^1_\indtype),\ldots,g(X^n_\indtype),w) = T.$$
  Finally, let $r =
  \bar{r}_{\worldtype\typearrow\worldtype\typearrow o}$. We choose $R_r$ such
  that $\langle w,w'\rangle\in R_{r}$ if and only if $(I
  r_{\worldtype\typearrow\worldtype\typearrow o})(w,w') = T$.
 \end{definition}

  It is not hard to verify that $Q^H=(W,(R_r)_{r\in S},D,P,(I_w)_{w\in
    W})$ meets the definition of $\QKPIm$ models. Below we will see that it
  also meets the definition of $\QKPI$ models.



 \begin{lemma} \label{lemma2} Let $Q^H=(W,(R_r)_{r\in S},D,P,(I_w)_{w\in W})$ be a $\QKPIm$ model
   for a given Henkin model $H=\langle
   \{D_\alpha\}_{\alpha\in\mathcal{T}}, I \rangle$. Furthermore, let $s = \bar{s}_{\worldtype\typearrow o}$.

 For all
   worlds $w\in W$ and variable assignments $g$ we have
   $\mathcal{V}_{[w/W_\worldtype]g}\, (s_{\worldtype\typearrow o} W_\worldtype) = T$
   in $H$ if and only if $Q^H,\bar{g},w \models s$ in $Q^H$.
\end{lemma}

\textbf{Proof:}
The proof is by induction on the structure of 
$s_{\worldtype\typearrow o}\in L$ and it is analogous to the proof of Lemma \ref{lemma1}.

$\Box$

With the help of Lemma \ref{lemma2}, we now show that the  $\QKPIm$ models we construct
in Def.\,\ref{def3} are in fact always  $\QKPI$ models. Thus, Henkin models 
never relate to $\QKPIm$ models that do not already fulfill the  $\QKPI$ criterion. 

\begin{lemma}\label{nominusmodels}
  Let $Q^H=(W,(R_r)_{r\in S},D,P,(I_w)_{w\in W})$ be a $\QKPIm$ model
   for a given Henkin model $H=\langle
   \{D_\alpha\}_{\alpha\in\mathcal{T}}, I \rangle$. Then $Q^H$ is also a $\QKPI$ model.
\end{lemma}

\textbf{Proof:} 
  We need to show that for every variable assignment $\bar{g}$ and
  formula $s=\bar{s}_{\worldtype\typearrow o}$ the set $\{w\in W \mid
  Q^h,\bar{g},w \models s\}$ is a member of $P$ in $Q^H$. This is a consequence
  of the 
\edcorr
{``Denotatpflicht''}
{rule that everything denotes}
in the Henkin model $H$. To see this, consider
  $\mathcal{V}_{g} s_{\worldtype\typearrow o} = \mathcal{V}_{g}
  (\lam{V_\worldtype} s_{\worldtype\typearrow o}\,V)$ for variable $V_\worldtype$ not occurring free in $s_{\worldtype\typearrow o}$. By definition of Henkin models this denotes 
  that function from $D_\worldtype=W$ to truth values $D_o=\{T,F\}$
  whose value for each argument $w\in D_\worldtype$ is
  $\mathcal{V}_{[w/V_\worldtype]g} (s\,V)$, that is, $s_{\worldtype\typearrow o}$ denotes the characteristic
  function $\lam{w\in W}
  \mathcal{V}_{[w/V_\worldtype]g}\, (s_{\worldtype\typearrow o} V_\worldtype) = T$
which we identify with the set $\{w\in W \mid
  \mathcal{V}_{[w/V_\worldtype]g}\, (s_{\worldtype\typearrow o} V_\worldtype) = T\}$.
  Hence, we have $\{w\in W \mid \mathcal{V}_{[w/V_\worldtype]g}\,
  (s_{\worldtype\typearrow o} V_\worldtype) = T\} \in D_{\worldtype\typearrow o}$.
  By the choice of $P=D_{\worldtype\typearrow o}$ in the construction of
  $Q^H$ we know $\{w\in W \mid
  \mathcal{V}_{[w/V_\worldtype]g}\, (s_{\worldtype\typearrow o} V_\worldtype) = T\}
  \in P$. By Lemma \ref{lemma2} we get $\{w\in W \mid
  Q^h,\bar{g},w \models s\} \in P$.

$\Box$

\begin{theorem}[Completeness for $\QKPI$ models] \label{thm2} Let $s_{\worldtype\typearrow o}$
  be a $\QMLSTT$ proposition and let $s = \bar{s}_{\worldtype\typearrow o}$
  be the corresponding $\QML$ proposition. If $\models^{\QKPI} s$  then $\models^{\stt}
  (\text{valid}\, s_{\worldtype\typearrow o})$.
\end{theorem}

\textbf{Proof:}
  By contraposition, assume $\not\models^{\stt}
  (\text{valid}\, s_{\worldtype\typearrow o})$: there is a Henkin
  model $H=\langle \{D_\alpha\}_{\alpha\in\mathcal{T}}, I \rangle$ and
  a variables assignment $g$ such that $\mathcal{V}_{g}\,
  (\text{valid}\, s_{\worldtype\typearrow o}) = F$. Hence, for some world
  $w\in D_\worldtype$ we have $\mathcal{V}_{[w/W_\worldtype]g}\,
  (s_{\worldtype\typearrow o} W_\worldtype) = F$.  By Lemma \ref{lemma2} we then
  get $Q^H,\bar{g},w \not\models^{\QKPIm} s$ for $s=\bar{s}_{\worldtype\typearrow
    o}$ in $\QKPIm$ model $Q^H$ for $H$. By Lemma
  \ref{nominusmodels} we know that $Q^H$ is actually a $\QKPI$
  model. Hence, $\not\models^{\QKPI} s$.
$Box$




Our soundness and completeness results obviously also apply to fragments
of $\QML$ logics.

\begin{corollary}
The reduction of our embedding to propositional quantified multimodal logics (which only allow quantification over propositional variables)
is sound and complete.
\end{corollary}
\begin{corollary}
The reduction of our embedding to first-order multimodal logics (which only allow quantification over individual variables)
is sound and complete.
\end{corollary}
\begin{corollary}
The reduction of our embedding to propositional multimodal logics (no
quantification) is sound and complete.
\end{corollary}

\vspace{\topsep}\noindent

\vfill\vfill\vfill
\section{Applying the Embedding in Practice} \label{sec5} In this
section, we illustrate the practical benefits of our embedding with
the help of some simple experiments.  We employ off-the-shelf
automated higher theorem provers and model generators for simple type
theory to solve problems in quantified multimodal logic. Future work
includes the encoding of a whole library of problems for quantified
multimodal logics and the systematic evaluation of the strengths of
these provers to reason about them.

In our case studies, we have employed the simple type theory automated
reasoners LEO-II,  TPS
\cite{DBLP:journals/japll/AndrewsB06}, IsabelleM and IsabelleP\@.\footnote{IsabelleM is
  a model finder in Isabelle that has been made available in batch
  mode, while IsabelleP  applies a series of Isabelle proof tactics in batch mode.}  These systems are available online
via the SystemOnTPTP tool and they exploit the new TPTP infrastructure
for typed higher-order logic \cite{BRS08}.

The formalization of the modal operators (Def.\,\ref{def1})
and the notion of validity (Def.\,\ref{validity}) 
in THF syntax \cite{BRS08} is presented in Appendix
\ref{app1}. As secured by the theoretical results of this paper, these
few lines of definitions are all we need to make simple type
theory reasoners applicable to quantified multimodal logic.

If we call the theorem provers LE0-II and IsabelleP with
this file, then they try to find a refutation from these
equations: they try to  prove their inconsistency. As expected, none of the systems reports success.  
The model finder IsabelleM, however, answers in 0.6 seconds that a model
has been found. IsabelleM employs the SAT solver zChaff. 

When applying our systems to Example \ref{ex2}, we get the following
results (where $+/t$ represents that a proof has been found in $t$
seconds and $-/t$ reports that no proof has been found within $t$
seconds): IsabelleP: $+/1.0$, LEO-II: $+/0.0$, TPS: $+/0.3$. IsabelleM
does not find a model (this also holds for the examples below).

\vspace{\topsep}\noindent
We also tried the Barcan formula and its converse: 
\begin{align*}
BF:\quad  & \text{valid}\,\, (\mall{X_\indtype} \mball{r} (p_{\indtype\typearrow(\worldtype\typearrow o)}\, X)) \mimpl (\mball{r} \mall{X_\indtype} (p_{\indtype\typearrow(\worldtype\typearrow o)}\, X)) \\
BF^{-1}:\quad & \text{valid}\,\, (\mball{r} \mall{X_\indtype} (p_{\indtype\typearrow(\worldtype\typearrow o)}\, X)) \mimpl (\mall{X_\indtype} \mball{r} (p_{\indtype\typearrow(\worldtype\typearrow o)}\, X))
\end{align*}
The results for $BF$ and $BF^{-1}$
are IsabelleP: $+/0.7$, LEO-II and LEO-IIP: $+/0.0$, TPS: $+/0.2$.
This confirms that our first-order quantification is constant domain.

\vspace{\topsep}\noindent
The next example analyzes the equivalence of two quantified multimodal
logic formula schemes (which can be read as ``if it is possible for everything to be $P$, then everything is potentially $P$\,''):
\begin{align*}
& \all{R_{\iota\typearrow\worldtype\typearrow o}}\all{P_{\indtype\typearrow(\worldtype\typearrow o)}} \\
& \quad (\text{valid}\,\,\, (\mdexi{R} \mall{X_\indtype} (P\,X)) \mimpl (\mall{X_\indtype} \mdexi{R} (P\, X))) \\
& \qquad \Leftrightarrow \\
& \quad (\text{valid}\,\,\,  (\mexi{X_\indtype} \mball{R} (P\,X))  \mimpl  (\mball{R} \mexi{X_\indtype} (P\,X))) 
\end{align*}
The results are: IsabelleP: $+/2.0$, LEO-II: $+/0.0$,
TPS:$+/0.2$.

\vfill\pagebreak

\vspace{\topsep}\noindent
An interesting meta property is the correspondence between axiom
$$\text{valid}\,\, 
\edcorr
{\mall{P{\indtype\typearrow(\worldtype\typearrow o)}}}
{\mall{P_{\indtype\typearrow(\worldtype\typearrow o)}}}
(\mdexi{i}
\mball{j} P) \mimpl \mball{k} \mdexi{l} P$$ and the
$(i,j,k,l)$-confluence property: 
$$\all{A_\worldtype}
\all{B_\worldtype} \all{C_\worldtype} (((i\, A\, B) \wedge (k\, A\,
C)) \Rightarrow \exi{D_\worldtype} ((j\, B\, D) \wedge (l\, C\, D)))$$
The results are: IsabelleP: $+/3.7$, LEO-II: $+/0.3$,
TPS:$+/0.2$. The problem encoding is presented in Appendix \ref{app4}.

\vspace{\topsep}\noindent
Future work will investigate how well this approach scales for more challenging
problems. We therefore invite potential users to encode their problems in the THF syntax and to submit them to
the THF TPTP library.



\vfill\pagebreak

\section{Conclusion} 
\label{Conclusion}

We have presented a straightforward embedding of quantified multimodal
logics in simple type theory and we have shown that this embedding is
sound and complete for $\QKPI$ semantics. This entails further soundness and
completeness results of our embedding for fragments of quantified
multimodal logics. We have formally explored the natural
correspondence between $\QKPI$ models and Henkin models and we have
shown that the weaker $\QKPI^-$ models do not enjoy such a
correspondence.

Non-quantified and quantified (normal) multimodal logics can thus
be uniformly seen as natural fragments  of simple type theory and their semantics (except some
weak notions such as  $\QKPI^-$ models) 
can be studied from the perspective of the well understood semantics of simple type theory.
Vice versa, via our embedding we can characterize some computationally interesting fragments
of simple type theory, which in turn may lead to some powerful proof tactics for
higher-order proof assistants. 

Future work includes further extensions of our
embedding to also cover quantified hybrid logics
\cite{BlackburnMarx02,Brauner05} and full higher-order modal logics
\cite{Fitting02book,Muskens06}. A first suggestion in direction of higher-order modal logics has
already been made \cite{B9}. This proposal does however not yet
address intensionality aspects. However, combining this proposal with
non-extensional notions of models for simple type theory \cite{BBK04,Muskens07}
appears a promising direction.


\vfill\pagebreak

\bibliographystyle{plain} 

\cleardoublepage

\begin{appendix} 
\section{THF Formalization of Quantified Multi-Modal Logic in Simple Type Theory}\label{app1}
{\footnotesize
\begin{verbatim}
%---------------------------------------------------------------------
% File     : QML.ax
% Domain   : Quantified multimodal logic 
% Problems : 
% Version  : 
% English  : Embedding of quantified multimodal logic in
%            simple type theory
% Refs     :  
% Source   : Formalization in THF by C. Benzmueller  
% Names    : 
% Status   : 
% Rating   : 
% Syntax   : 
% Comments : 
%---------------------------------------------------------------------
%---- declaration of additional base type mu
thf(mu,type,(
    mu: $tType )).

%---- modal operators not, or, box, Pi (for types mu and $i>$o)
thf(mnot,definition,
    ( mnot
    = ( ^ [Phi: $i > $o,W: $i] :
          ~ ( Phi @ W ) ) )).

thf(mor,definition,
    ( mor
    = ( ^ [Phi: $i > $o,Psi: $i > $o,W: $i] :
          ( ( Phi @ W )
          | ( Psi @ W ) ) ) )).

thf(mbox,definition,
    ( mbox
    = ( ^ [R: $i > $i > $o,Phi: $i > $o,W: $i] :
        ! [V: $i] :
          ( ~ ( R @ W @ V )
          | ( Phi @ V ) ) ) )).

thf(mall_ind,definition,
    ( mall_ind
    = ( ^ [Phi: mu > $i > $o,W: $i] :
        ! [X: mu] :
          ( Phi @ X @ W ) ) )).

thf(mall_prop,definition,
    ( mall_prop
    = ( ^ [Phi: ( $i > $o ) > $i > $o,W: $i] :
        ! [P: $i > $o] :
          ( Phi @ P @ W ) ) )).

%---- further modal operators
thf(mtrue,definition,
    ( mtrue
    = ( mall_prop
      @ ^ [P: $i > $o] :
          ( mor @ P @ ( mnot @ P ) ) ) )).

thf(mtrue,definition,
    ( mfalse
    = ( mall_prop
      @ ^ [P: $i > $o] :
          ( mnot @ mtrue ) ) )).

thf(mand,definition,
    ( mand
    = ( ^ [Phi: $i > $o,Psi: $i > $o] :
          ( mnot @ ( mor @ ( mnot @ Phi ) @ ( mnot @ Psi ) ) ) ) )).

thf(mimpl,definition,
    ( mimpl
    = ( ^ [Phi: $i > $o,Psi: $i > $o] :
          ( mor @ ( mnot @ Phi ) @ Psi ) ) )).

thf(mdia,definition,
    ( mdia
    = ( ^ [R: $i > $i > $o,Phi: $i > $o] :
          ( mnot @ ( mbox @ R @ ( mnot @ Phi ) ) ) ) )).

thf(mexi_ind,definition,
    ( mexi_ind
    = ( ^ [Phi: mu > $i > $o] :
          ( mnot
          @ ( mall_ind
            @ ^ [X: mu] :
                ( mnot @ ( Phi @ X ) ) ) ) ) )).

thf(mexi_prop,definition,
    ( mexi_prop
    = ( ^ [Phi: ( $i > $o ) > $i > $o] :
          ( mnot
          @ ( mall_prop
            @ ^ [P: $i > $o] :
                ( mnot @ ( Phi @ P ) ) ) ) ) )).

%---- definition of validity
thf(mvalid,definition,
    ( mvalid
    = ( ^ [Phi: $i > $o] :
        ! [W: $i] :
          ( Phi @ W ) ) )).
\end{verbatim}
}

\vfill\pagebreak

\section{THF Example: In all Worlds exists Truth}  \label{app2}
{\footnotesize
\begin{verbatim}
%--------------------------------------------------------------------
% File     : ex1.p
% Domain   : Quantified multimodal logic 
% Problems : 
% Version  : 
% English  : In all accessible worlds exists truth.
% Refs     :  
% Source   : Formalization in THF by C. Benzmueller  
% Names    : 
% Status   : 
% Rating   : 
% Syntax   : 
% Comments : 
%--------------------------------------------------------------------
%---- include the definitions for qunatified multimodal logic 
include('QML.ax').

%---- provide a consant for accesibility relation r
thf(r,type,r:$i>$i>$o).

%---- conjecture statement
thf(ex1,conjecture,
    (mvalid @ (mbox @ r @ (mexi_prop @ (^[P:$i>$o]: P))))).
\end{verbatim}
}


\vfill\pagebreak

\section{THF Example:  Confluence Property of Accessibility Relations}  \label{app4}
{\footnotesize
\begin{verbatim}
%--------------------------------------------------------------------
% File     : ex9.p 
% Domain   : Quantified multimodal logic 
% Problems : 
% Version  : 
% English  : Confluence property of accessibility relations
% Refs     :  
% Source   : Formalization in THF by C. Benzmueller  
% Names    : 
% Status   : 
% Rating   : 
% Syntax   : 
% Comments : 
%--------------------------------------------------------------------
%---- include the definitions for qunatified multimodal logic 
include('QML.ax').

%---- constants for accesibility relations
thf(i,type,(
    i: $i > $i > $o )).

thf(j,type,(
    j: $i > $i > $o )).

thf(k,type,(
    k: $i > $i > $o )).

thf(l,type,(
    l: $i > $i > $o )).

%---- definition of confluence property   
thf(confluence,definition,
    ( confluence
    = ( ^ [I: $i > $i > $o,J: $i > $i > $o,
           K:  $i > $i > $o,L: $i > $i > $o] :
        ! [A: $i,B: $i,C: $i] :
          ( ( ( I @ A @ B )
            & ( K @ A @ C ) )
         => ? [D: $i] :
              ( ( J @ B @ D )
              & ( L @ C @ D ) ) ) ) )).

%---- correspondence between axiom and confluence property  
thf(conj,conjecture,
    ( ( mvalid
      @ ( mall_prop
        @ ^ [P: $i > $o] :
            ( mimpl @ ( mdia @ i @ ( mbox @ j @ P ) )
                    @ ( mbox @ k @ ( mdia @ l @ P ) ) ) ) )
  <=> ( confluence @ i @ j @ k @ l ) )).

\end{verbatim}
}

\end{appendix}

\end{document}